Receptor System. Front Pharmacol. 13, 856672. [2] Leconte, C., Patricia Sales, A., Bergoin, E., Beray-Berthat, V., Noble, F., Mongeau, R. 2020. Traumatic-like fear memory recall causes persistent morphine seeking behavior in mice. European Neuropsychopharmacol. 40(1), S3-S4. [3] Daumas, S., Betourne, A., Halley, H., Wolfer, D.P., Lipp, H-P., Lassalle, J-M., Francés, B. 2007. Transient activation of the CA3 Kappa opioid system in the dorsal hippocampus modulates complex memory processing in mice. Neurobiol Learn Mem. 88(1), 94-103.
No conflict of interest

doi: https://doi.org/10.1016/j.nsa.2023.102833

## P.0026
### NEUROSCIENCE APPLIED 2 (2023) 102439 102834
### GROUP-BASED SURFACE STRUCTURAL COVARIANCE ANALYSIS OF ADHD IN THE ENIGMA DATASET

O. Grimm [1], Z. Mossawaty [2], E.A.G. Please change [3], B. Franke [4], M. Hoogman [5]. [1] *Goethe-University-Frankfurt, University Clinic- Departement of Psychiatry-Psychotherapy and Psychosomatics, Frankfurt Am Main, Germany;* [2] *Goethe University, University Clinic- Departement Psychiatry- Psychosomatics and Psychotherapy, Frankfurt, Germany;* [3] *Please change, Please change, Please change, Germany;* [4] *Radboud University, Radboud Medical centre, Nijmegen, Netherlands;* [5] *Radboud University, Radboud University Medical Centre Nijmegen, Nijmegen, Netherlands*

Attention Deficit Hyperactivity Disorder (ADHD) is a neurodevelopmental disorder that has been linked to delayed brain maturation, particularly in early ADHD cases. Structural Magnetic Resonance Imaging (MRI) can provide an understanding of how cortical thickness and surface area behave in relation to ADHD diagnosis. However, developmental patterns are not easily understood using region-of-interest analyses. A recent study of the ENIGMA consortium demonstrated lower surface area in children [1]. The technique of structural covariance can be used to understand how different brain regions are related to each other in terms of morphological properties at the group level [2].
We conducted a cortical structural covariance analysis of the ADHD ENIGMA mega-analysis dataset, which consisted of over 3000 MRI datasets in both children and adults, with more than 1500 ADHD cases. To investigate the alterations in the architecture of structural networks in LLD compared with controls, we applied graph theoretical methods using the GAT [3], which integrates the Brain Connectivity Toolbox [4] for the calculation and statistical comparisons of network measures. Specifically, networks were constructed for the ADHD and control group for both children as well as adults separately using the structural covariance approach). The nodes in the network correspond to the cortical FreeSurfer parcellations. Consistent with previous studies, linear regression was performed at each ROI to remove the effects of covariates, including age, site, and mean intracranial volume The resulting residuals of this regression are then substituted for the unadjusted cortical measures at each ROI. Comparison between correlation matrices from the parcellations were done for the ADHD and the control groups and comparison was done with a permutation-based statistic. The results showed that there was a higher cumulative degree distribution in ADHD compared to healthy controls (HC) in terms of surface-based structural covariance. At a network matrix density between 0.25 and 0.3, the characteristic path length was also higher in ADHD. When the researchers looked at regional hubs driving these effects, they found that the normalized degree was significantly higher in ADHD ($p=0.001$), whereas the rostral Anterior Cingulate Cortex (ACC) was significantly better connected in HC ($p=0.036$). These effects were not found in adults, and to a lesser degree in the analysis of cortical thickness covariance.
In summary, the results suggest that surface structural covariance is higher in children below the age of 16 across 17 different sites from the ADHD ENIGMA consortium. This points to a higher degree but less functional segregation in higher cortical areas. These results fit well with the developmental delay hypothesis and question a uniform, gradual pathophysiology between children and adults.
Overall, this study provides evidence for the use of structural covariance analysis in understanding ADHD at a group level. It highlights the importance of studying developmental patterns in ADHD and emphasizes the need for further research to explore the differences between children and adults with ADHD.

### References
1.Hoogman, M. et al. Brain Imaging of the Cortex in ADHD: A Coordinated Analysis of Large-Scale Clinical and Population-Based Samples. Am J Psychiat 176, 531–542 (2019). 2.Alexander-Bloch, A., Raznahan, A., Bullmore, E. & Giedd, J. The convergence of maturational change and structural covariance in human cortical networks. J Neurosci 33, 2889–2899 (2013). 3.Hosseini, S. M. H., Hoeft, F. & Kesler, S. R. GAT: a graph-theoretical analysis toolbox for analyzing between-group differences in large-scale structural and functional brain networks. Plos One 7, e40709 (2012). 4.Rubinov, M. & Sporns, O. Complex network measures of brain connectivity: Uses and interpretations. Neuroimage 52, 1059–1069 (2010).
No conflict of interest

doi: https://doi.org/10.1016/j.nsa.2023.102834

## P.0027
### NEUROSCIENCE APPLIED 2 (2023) 102439 102835
### ADHD DIAGNOSIS BASED ON ACTION CHARACTERISTICS RECORDED IN VIDEOS USING MACHINE LEARNING

Y. Li [1], S. Mohsen Naqvi [1], R. Nair [2]. [1] *Newcastle University, Intelligent sensing and Communications research group, Newcastle, United Kingdom;* [2] *NHS, Adult ADHD SERVICE, GATESHEAD, United Kingdom*

**Introduction:** Demand for ADHD diagnosis and treatment is increasing significantly and the existing services are unable to meet the demand in a timely manner. In this work, we introduce a novel action recognition method for ADHD diagnosis by identifying and analysing raw video recordings. Our main contributions include 1) designing and implementing a test focusing on the attention and hyperactivity/impulsivity of participants, recorded through three cameras; 2) implementing a novel machine learning ADHD diagnosis system based on action recognition neural networks for the first time; 3) proposing classification criteria to provide diagnosis results and analysis of ADHD action characteristics.
**Aim of the study:** We propose a novel ADHD diagnosis system with an action recognition framework, utilizing a multi-modal ADHD dataset and state-of-the-art machine learning detection algorithms. Our proposed method can perform a visual analysis of the participants' action characteristics, which can be combined with clinical information to enhance the accuracy and comprehensiveness of the assessment. Compared to conventional **methods:** the proposed method shows cost-efficiency and significant performance improvement, making it more accessible for a broad range of initial ADHD diagnoses.
**Statistical analysis methods:** The dataset contains 7 ADHD patients over 18years of age and 10 controls provided by the CNTW-NHS Foundation Trust and Newcastle University. Data was captured in the intelligent sensing laboratory at Newcastle University when they were interviewed using a questionnaire based on DSM V and completed cognitive tasks. The actions of subjects and controls in our dataset mainly contain three categories: still position, limb fidgets, and torso movements. State-of-the-art machine learning algorithms recognize and label the actions in the test videos. We propose a novel Hyperactivity Score (HS) and a measurement named Attention Deficit Ratio (ADR) as the evaluation criterion for action classification of ADHD symptoms detection. They focus on the action change frequency of the subjects and controls during the test, which is also defined as the model's ability to focus on the movement or posture.
The performance of the proposed ADHD diagnosis system is evaluated by the standard measurements, e.g., accuracy, F1 score, precision, and the area under curve (AUC).
**Results:** In the result part, we compare the diagnostic result of our proposed method framework with five typical action recognition neural networks: C3D, R3D, MS-G3D, ST-GCN, and PoseC3. The results are shown in Table 1:

|          | Precision | F1 Score | Accuracy | AUC  |
| -------- | --------- | -------- | -------- | ---- |
| R3D [1]  | 58.8      | 74.0     | 58,8     | 0.50 |
| C3D [2]  | 85.7      | 70.6     | 70.6     | 0.71 |
| ST-GCN [3] | 100.0   | 75.0     | 76.4     | 0.70 |
| MS-G3D [4] | 85.7    | 70.6     | 70.6     | 0.72 |
| PoseC3D [5] | 100.0  | 88.9     | 88.2     | 0.83 |

**Conclusion**: Our systems are cost-effective and easily integrable into clinical practice. Furthermore, we will focus on the effectiveness of deep learning models, particularly those based on graph convolutional networks and spatial-temporal architectures, to achieve superior results in action recognition task, thereby enabling the development of more efficient diagnostic systems for various applications. This includes implementing the AI in portable devices like mobile phones, laptops etc, which would help with early, accurate diagnosis.
### References
[1] D. Tran, L. Bourdev, R. Fergus, L. Torresani, and M. Paluri.,2015. "Learning

spatiotemporal features with 3D convolutional networks," in Proceedings of the IEEE international conference on computer vision (ICCV). [2] D. Tran, H. Wang, L. Torresani, J. Ray, Y. LeCun, and M. Paluri.,2018. "A closer look at spatiotemporal convolutions for action recognition," in Proceedings of the IEEE conference on Computer Vision and Pattern Recognition (CVPR). [3] K. Cheng, Y. Zhang, X. He, W. Chen, J. Cheng, and H Lu.,2020. "Skeleton-based action recognition with shift graph convolutional network," in Proceedings of the IEEE conference on Computer Vision and Pattern Recognition (CVPR). [4] Z. Liu, H. Zhang, Z. Chen, Z. Wang, and W. Ouyang.,2020. "Disentangling and unifying graph convolutions for skeleton-based action recognition," in Proceedings of the IEEE conference on Computer Vision and Pattern Recognition (CVPR). [5] H. Duan, Y Zhao, K. Chen, D. Lin, and B Dai.,2022. "Revisiting skeleton-based action recognition," in Proceedings of the IEEE Conference on Computer Vision and Pattern Recognition (CVPR).

**Conflict of interest:**
**Disclosure statement:** I HAVE RECIVED HONORARIA FROM TAKEDA AND LUNDBECK, but not for this study.
I have received a small grant from Newcastle university to conduct the study.

doi: https://doi.org/10.1016/j.nsa.2023.102835

P.0029
NEUROSCIENCE APPLIED 2 (2023) 102439 102836
EVALUATING THE USE OF PHARMACOLOGICAL TREATMENTS FOR ATTENTION-DEFICIT HYPERACTIVITY DISORDER IN AUTISTIC INDIVIDUALS: A RETROSPECTIVE COHORT STUDY

G. Gillett [1], H. Hopkins [1], N. Adamo [1]. [1] *King's College London, Institute of Psychiatry- Psychology & Neuroscience, London, United Kingdom*

**Introduction:** Attention-deficit hyperactivity disorder (ADHD) and Autism Spectrum Disorder (ASD) are common co-occurring conditions. Clinical guidance suggests that ADHD medications should be prescribed for individuals with co-occurring ADHD and ASD, akin to recommendations for individuals diagnosed with ADHD alone. However, previous research suggests that individuals with ADHD and ASD are prescribed ADHD medications less frequently compared to individuals with ADHD alone, and that lower doses are typically used, possibly reflecting differences in clinical effectiveness, acceptability or tolerability [1]. This hypothesis is supported by results from a number of small studies demonstrating poor tolerability and acceptability among autistic individuals taking methylphenidate [2], although evidence is inconsistent and limited [3, 4].
**Methods:** We analysed the clinical records of patients treated in South London and Maudsley Trust from commencement of the electronic records system until the present day. Using the Clinical Record Interactive Search (CRIS) system, natural language processing was applied to identify individuals diagnosed with both ASD and ADHD, as well as a control cohort of individuals diagnosed with ADHD alone [5]. Natural language processing was also used to extract the dates of any prescribed medication for ADHD, standardised clinical outcomes and free-text descriptions of common adverse effects, alongside demographic data.
Statistical analyses compared demographic characteristics, baseline clinical data and the number, type and duration of prescribed medications between ADHD and ASD/ADHD cohorts. In further analyses, Cox proportional hazards regression will be used to calculate the hazard ratios for medication discontinuation and occurrence of specific adverse effects for each medication, between the ADHD and ASD/ADHD cohorts. Standardised clinical outcomes will also be compared between the ADHD and ASD/ADHD cohorts as proxy measures of clinical effectiveness.
**Results:** This poster presentation will report preliminary results from this retrospective cohort study. The following results will be presented: i) the baseline clinical and demographic characteristics of each cohort; ii) the number of prescribed ADHD medications and the duration over which they were taken for each cohort; iii) prescribing rates of ADHD medication by specific medication and type (stimulant vs non-stimulant) for each cohort; and; iv) the sequence in which ADHD medications were trialled for each cohort. Results from statistical analyses comparing ADHD and ASD/ADHD cohorts will be presented for each domain.
**Conclusion:** Our findings contribute to a small, emerging literature assessing possible differences in clinical prescribing patterns of ADHD medications in autistic individuals. In particular, our large, externally valid, observational clinical dataset may provide insights not readily available from other study designs. Our results may have direct implications for clinicians working in child, adolescent and adult services, as well as informing future research directions in this field. Future work will further characterise outcomes relating to clinical effectiveness, acceptability and tolerability among autistic individuals prescribed ADHD medication.

**References**
1. Johansson V, Sandin S, Chang Z, et al. Medications for attention-deficit/hyperactivity disorder in individuals with or without coexisting autism spectrum disorder: analysis of data from the Swedish prescribed drug register. J Neurodev Disord. 2020;12(1):44. 2. Rodrigues R, Lai MC, Beswick A, Gorman DA, Anagnostou E, Szatmari P, Anderson KK, Ameis SH. Practitioner Review: Pharmacological treatment of attention-deficit/hyperactivity disorder symptoms in children and youth with autism spectrum disorder: a systematic review and meta-analysis. J Child Psychol Psychiatry. 2021 Jun;62(6):680-700. 3. Joshi G, Wilens T, Firmin ES, Hoskova B, Biederman J. Pharmacotherapy of attention deficit/hyperactivity disorder in individuals with autism spectrum disorder: A systematic review of the literature. J Psychopharmacol. 2021;35(3):203-210. 4. Muit JJ, Bothof N, Kan CC. Pharmacotherapy of ADHD in Adults With Autism Spectrum Disorder: Effectiveness and Side Effects. J Atten Disord. 2020;24(2):215-225. 5. Perera G, Broadbent M, Callard F, et al. Cohort profile of the South London and Maudsley NHS Foundation Trust Biomedical Research Centre (SLaM BRC) Case Register: current status and recent enhancement of an Electronic Mental Health Record-derived data resource. BMJ Open. 2016;6(3):e008721.
No conflict of interest

doi: https://doi.org/10.1016/j.nsa.2023.102836

P.0030
NEUROSCIENCE APPLIED 2 (2023) 102439 102837
SALIENCE AND HEDONIC EXPERIENCE AS PREDICTORS OF CENTRAL STIMULANT TREATMENT RESPONSE IN ADHD

J. Rode [1,2], R. Runnamo [3], P. Thunberg [1,4], M. Msghina [3,5]. [1] *Faculty of Medicine and Health- Örebro University, Center for Experimental and Biomedical Imaging in Örebro CEBIO, Örebro, Sweden;* [2] *Örebro University- School of Medical Sciences, Nutrition-Gut-Brain Interactions Research Centre- Faculty of Medicine and Health, Örebro, Sweden;* [3] *Faculty of Medicine and Health- Örebro University, Department of Psychiatry- School of Medical Sciences, Örebro, Sweden;* [4] *Örebro University, Department for Radiology and Medical Physics- Faculty of Medicine and Health, Örebro, Sweden;* [5] *Karolinska Institute, Department of Clinical Neuroscience, Stockholm, Sweden*

**Background:** Attention-deficit/hyperactivity disorder (ADHD) is a common multifactorial neurodevelopmental disorder with an often profound and pervasive effect on the individual. ADHD is associated with multiple psychiatric comorbidities such as depression [1], anxiety disorder [1] and substance abuse [1]; and ADHD patients generally experience numerous difficulties in life, such as having a higher risk of not completing higher education and being unemployed [2]. Standard treatment for ADHD is multimodal, with central stimulants (CS) largely considered the backbone of the treatment. From previous studies, it is known that approximately 20-30% of patients with ADHD are non-responsive to CS medication [3]. Despite multiple studies investigating biochemical, neuroimaging, genetic, and behavioral biomarkers, there are no clinically useful biomarkers identified, that can help distinguish CS responders and non-responders.
**Methods:** We investigated if incentive salience and hedonic experience assessed after a single-dose CS medication could predict response and non-response to CS medication in ADHD patients. In order to do this, we utilized a bipolar visual analogue 'wanting' and 'liking' scale evaluating incentive salience and hedonic experience in 29 patients diagnosed with ADHD and 25 healthy controls. Healthy controls were administered 30 mg methylphenidate and ADHD patients received either methylphenidate or lisdexamphetamine in individual dosage in order to optimize effect, as selected by their clinical psychiatrist.
Clinician evaluated global impression improvement (CGI-I), was utilized to evaluate response to CS medication. Resting state functional magnetic resonance imaging (fMRI) was performed, using a 3.0T MR system and 48-channel head coil, before and after single-dose CS to assess changes in functional connectivity. We also stratified based on the covariates 'liking' and 'wanting', using this as an experimental definition of response.
**Results:** Approximately 20% of ADHD patients in the study were non-responders to CS medication (5 of 29). Responders had significantly higher incentive salience and hedonic experience scores compared to healthy controls and non-responders. Resting state fMRI showed that wanting scores were significantly associated to changes in functional connectivity in ventral striatum including nucleus accumbens, areas dense in dopaminergic innervation and part of the brain reward system.